\def\year{2022}\relax
\documentclass[letterpaper]{article} 
\usepackage{aaai22}  
\usepackage{times}  
\usepackage{helvet}  
\usepackage{courier}  
\usepackage[hyphens]{url}  
\usepackage{graphicx} 
\usepackage{natbib}  
\usepackage{caption} 
\DeclareCaptionStyle{ruled}{labelfont=normalfont,labelsep=colon,strut=off} 
\usepackage{algorithm}
\usepackage{algorithmic}

\usepackage{amsmath}
\usepackage{amssymb}
\usepackage{subcaption}
\usepackage{booktabs}       
\usepackage{amsfonts}       
\usepackage{nicefrac}       
\usepackage{microtype}      
\usepackage{xcolor}         
\input{preamble}

\usepackage{newfloat}
\usepackage{listings}
\floatstyle{ruled}
\newfloat{listing}{tb}{lst}{}
\floatname{listing}{Listing}
\title{Competing Mutual Information Constraints with Stochastic Competition-based Activations for Learning Diversified Representations}
\author{
	Konstantinos P. Panousis\textsuperscript{\rm 1}\footnote{Both authors contributed equally.}, 
	Anastasios Antoniadis\textsuperscript{\rm 1$*$}, 
	Sotirios Chatzis\textsuperscript{\rm 1}
}
\affiliations{
	\textsuperscript{\rm 1}Department of Electrical Engineering and Computer Engineering and Informatics \\
	Cyprus University of Technology, Limassol, Cyprus\\
	k.panousis@cut.ac.cy
}

\begin{document}
	
	\maketitle
	
	\begin{abstract}
		This work aims to address the long-established problem of learning diversified representations. To this end, we combine information-theoretic arguments with stochastic competition-based activations, namely Stochastic Local Winner-Takes-All (LWTA) units. In this context, we ditch the conventional deep architectures commonly used in Representation Learning, that rely on non-linear activations; instead, we replace them with sets of locally and stochastically competing linear units. In this setting, each network layer yields sparse outputs, determined by the outcome of the competition between units that are organized into blocks of competitors. We adopt stochastic arguments for the competition mechanism, which perform posterior sampling to determine the winner of each block. We further endow the considered networks with the ability to infer the sub-part of the network that is essential for modeling the data at hand; we impose appropriate stick-breaking priors to this end. To further enrich the information of the emerging representations, we resort to information-theoretic principles, namely the Information Competing Process (ICP). Then, all the components are tied together under the stochastic Variational Bayes framework for inference. We perform a thorough experimental investigation for our approach using benchmark datasets on image classification. As we experimentally show, the resulting networks yield significant discriminative representation learning abilities. In addition, the introduced paradigm allows for a principled investigation mechanism of the emerging intermediate network representations. 
	\end{abstract}

	%
	%
	%
	\section{Introduction}
	The ability to extract diversified representations is the main focal point of Representation Learning (RL). Despite the immense amount of research effort, though, learning representations with such qualities remains an open research question. Recently, information-theoretic arguments, and specifically mutual information constraints \cite{bell,belghazi2018mine,dim,contrastive}, have risen as a promising direction towards successful RL. Notably, the Information Bottleneck (IB) \cite{tishby99information} has attracted a substantial amount of attention \cite{alemi2017,achille2017information,dai2018compressing}. The main IB principle lies on constraining the inferred representations, such that the information carried about the input is minimized, while being maximally informative about the target outputs \cite{tishby15}. The first implementations of the IB principle in deep learning can be found in \cite{ShwartzZiv2017OpeningTB,alemi2017, michael2018on}. 
	
	Despite these advances, it is striking that the majority of the proposed works refrain from examining a core -and probably the most important- aspect of modern deep architectures; that is, the employed non-linearities. We posit that a radically different paradigm of latent unit operation may allow for significantly enhancing the representation power of deep networks. We draw inspiration from biologically-plausible architectures: it has been shown that neurons with similar functions in the mammal brain aggregate together in groups/blocks, and a local competition takes place for their activation \citep{kandel2000principles,andersen1969participation,stefanis1969interneuronal,douglas2004neuronal,lansner2009associative}. This leads to a Local Winner-Takes-All (LWTA) mechanism; the winner of the competition in each block gets to convey its activation outside the block, while the rest are inhibited to silence. 
	The incorporation of the LWTA mechanism in deep architectures has been shown to exhibit significant properties such as \textit{noise suppression, robustness to adversarial attacks} and \textit{compression} \cite{srivastava2013compete,grossberg1982contour,carpenter1988art,panousis2019nonparametric,panousisnips,panousis2021}. Despite these properties, LWTA-based networks have been scarcely examined in the RL literature. 
	
	On this basis, in this work, we propose a novel deep learning design framework that: (i) replaces the commonly employed non-linear activations with \textit{stochastic} LWTA activations; (ii) endows the considered networks with a data-driven c\textit{omponent utility inference mechanism}, which allows to infer the \textit{essential} sub-parts of the network necessary for modeling the available data; and (iii) utilizes recent advances in the information-theoretic approaches to yield diversified representations, namely \textit{competing mutual information constraints} \cite{icp}.
	Our ultimate goal is to yield deep networks with considerably diversified resulting representations. We evaluate our approach using well-known benchmark datasets and architectures for image classification.
	
	The rest of the paper is organized as follows: In Section \ref{sec:lwta}, we introduce the foundation of our methodology. In Section~\ref{sec:background}, we formulate our novel representation method, and provide associated training and prediction algorithms. In Section \ref{sec:experimental}, we perform rigorous experimental evaluations, investigate the behavior of the proposed framework, and provide strong separability results and insights into the emerging representations. Finally, in Section~\ref{sec:conclusions}, we provide some final remarks.

	\section{Foundational Principles}
	\label{sec:lwta}
	
	Let us assume a dense layer of a conventional deep neural network comprising $K$ hidden units. When presented with an input $\boldsymbol x \in \mathbb{R}^J$, each hidden unit $k$ performs an inner product computation, $h_k = \boldsymbol w_k^T \boldsymbol x = \sum_{j=1}^J w_{jk} \cdot x_j \in \mathbb{R}$; $\boldsymbol W \in \mathbb{R}^{J\times K}$ is the associated weight matrix of the layer. This \textit{response} usually passes through a non-linear function $\sigma(\cdot)$, yielding $y_k = \sigma(h_k), \ \forall k$. Thus, the final output of the layer arises from the concatenation of the non-linear \textit{activations} of each unit, such that $\boldsymbol y = [y_1, \dots, y_K] \in \mathbb{R}^K$.
	
	\begin{figure*}[t!]
		\centering
		\includegraphics[scale=0.93]{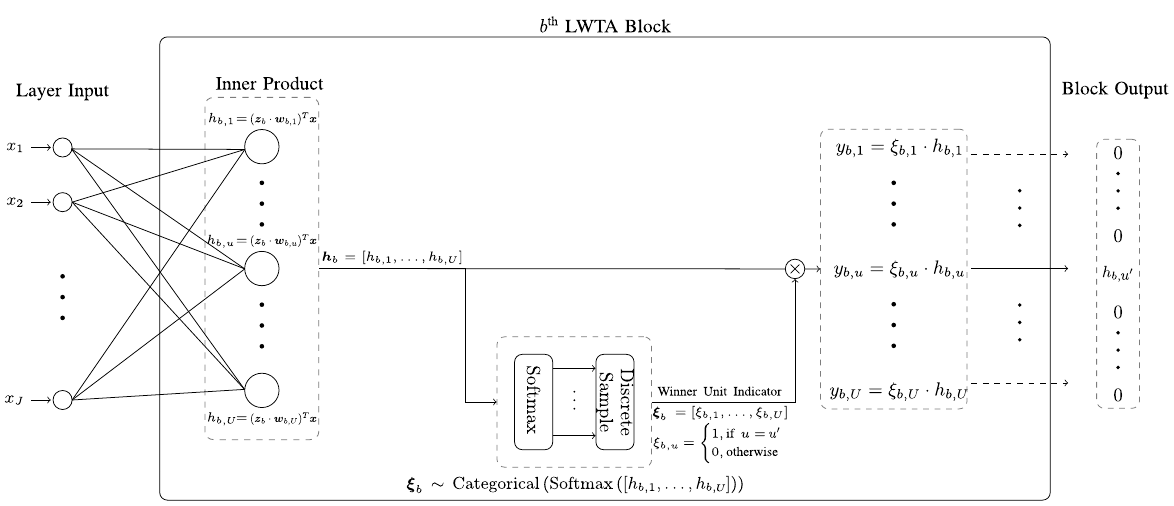}
		\caption{A detailed bisection of the $b$\textsuperscript{th} Stochastic LWTA block in an LWTA layer. Presented with an input $\boldsymbol x\in \mathbb{R}^J$,  each unit $u=1, \dots,U$ computes its activation $h_{b,u}$ via different weights $\boldsymbol w_{b,u}\in \mathbb{R}^J$, i.e., $h_{b,u} = (\boldsymbol{z}_b \cdot \boldsymbol{w}_{b,u}^T)\boldsymbol x$. Here, $\boldsymbol{z}_b$ is the component utility indicator pertaining to the $b$th block, which encodes which synapses leading to the $b$th block the inference algorithm deems useful, and which not. The linear responses of the units are concatenated, such that  $\boldsymbol h_b = [h_{b,1}, \dots, h_{b,U}]$, and transformed into probabilities via the softmax operation. Then, a Discrete sample $\boldsymbol \xi_b = [\xi_{b,1},\dots, \xi_{b,U}]$ is drawn; this constitutes a one-hot vector with a single non-zero entry at position $u'$, denoting the winner unit in the block. The winner unit passes its linear response to the next layer; the rest pass zero values.
		}
		\label{fig:block}
	\end{figure*}
	
	In contrast, in the LWTA framework, singular nonlinear hidden units are replaced by $U$ \textit{competing linear units} grouped together in an (LWTA) block; each layer comprises multiple such blocks. Hereinafter, we denote by $B$ the number of such blocks in a particular LWTA-based layer. The weight synapses are now structured as a three dimensional matrix $\boldsymbol W \in \mathbb{R}^{J \times B \times U}$, revealing that, in this case, each input is now presented to each block $b$ and each unit $u$ therein. In this context, each linear unit $u$ in each block $b$ computes its response, following the conventional inner product computation, such that $h_{b,u} = \boldsymbol w_{b,u}^T \boldsymbol x = \sum_{j=1}^j w_{j,b,u} \cdot x_j \in \mathbb{R}$; then, competition takes places among the units in the block.
	
	The main operating principle is that out of the $U$ units in the block, \textit{only one} can be \textit{the winner}; this unit gets to convey its (linear) activation to the next layer, while  \textit{all the rest are inhibited to silence}, i.e. pass a zero value. The final output of an LWTA-based layer $\boldsymbol y \in \mathbb{R}^{B \cdot U}$ is now composed of $B$ subvectors $\boldsymbol y_b \in \mathbb{R}^U$, one for each LWTA block and each with a \textit{single non-zero entry}. It is evident that this process results in a \textit{sparse representation}; in each block, only one out of the $U$ units produces a non-zero output{\protect\footnote{The higher the number of competitors $U$ in each block, the sparser the output. When $U=2$, only $50\%$ of the units in a layer are active for each example, when $U=4$, only $25\%$, e.t.c.}}. In the related LWTA literature, the competition process is usually deterministic, i.e., the winner with the \textit{highest linear activation} is deemed the winner each time. However, novel data-driven stochastic arguments for the competition process have been recently proposed in \citep{panousis2019nonparametric, panousisnips, panousis2021, Voskou_2021_ICCV}.
	
	In this setting, to encode the outcome of the competition in each of the $B$ stochastic LWTA blocks that constitute a stochastic LWTA layer, we introduce an appropriate set of discrete latent vectors $\boldsymbol \xi \in \mathrm{one\_hot}(U)^B$. This vector comprises $B$ component subvectors; each component entails \textit{exactly one non-zero value} at the \textit{index position} that corresponds to the \textit{winner unit} in each respective LWTA block. 
	
	Further, in this work, we introduce an additional data-driven mechanism in order to endow the networks with the ability to \textit{infer} which subparts of the network are \textit{essential} for modeling the data at hand. To this end, we introduce a matrix of auxiliary binary latent variables $\boldsymbol Z \in \{0,1 \}^{J \times B}$. Each entry therein is $1$ if the $j$\textsuperscript{th} component of the input is \textit{presented} to the $b$\textsuperscript{th} block and zero otherwise. In this setting, the response of each unit is facilitated via the inner product operation between the \textit{effective} network weights (as dictated by the latent indicators $\boldsymbol Z$) and the input.
	
	We can now express the output $\boldsymbol y$ of a stochastic LWTA layer's $(b,u)$\textsuperscript{th} unit, $y_{b,u}$, that is, the output of the $u$\textsuperscript{th} unit in the $b$\textsuperscript{th} block, as:
	\begin{align}
		y_{b,u} = \xi_{b,u} \sum_{j=1}^J (z_{j,b} \cdot w_{j,b,u}) \cdot x_j \in \mathbb{R}
		\label{eqn:dense_y}
	\end{align}
	where $\xi_{b,u}$ denotes the $u$\textsuperscript{th} component of $\boldsymbol \xi_b$, and $\boldsymbol \xi_b \in \mathrm{one\_hot}(U)$ holds the $b$\textsuperscript{th} subvector of $\boldsymbol \xi$.

\begin{figure*}
		\begin{subfigure}[t]{.44\textwidth}
			\centering
			\includegraphics[scale=0.5]{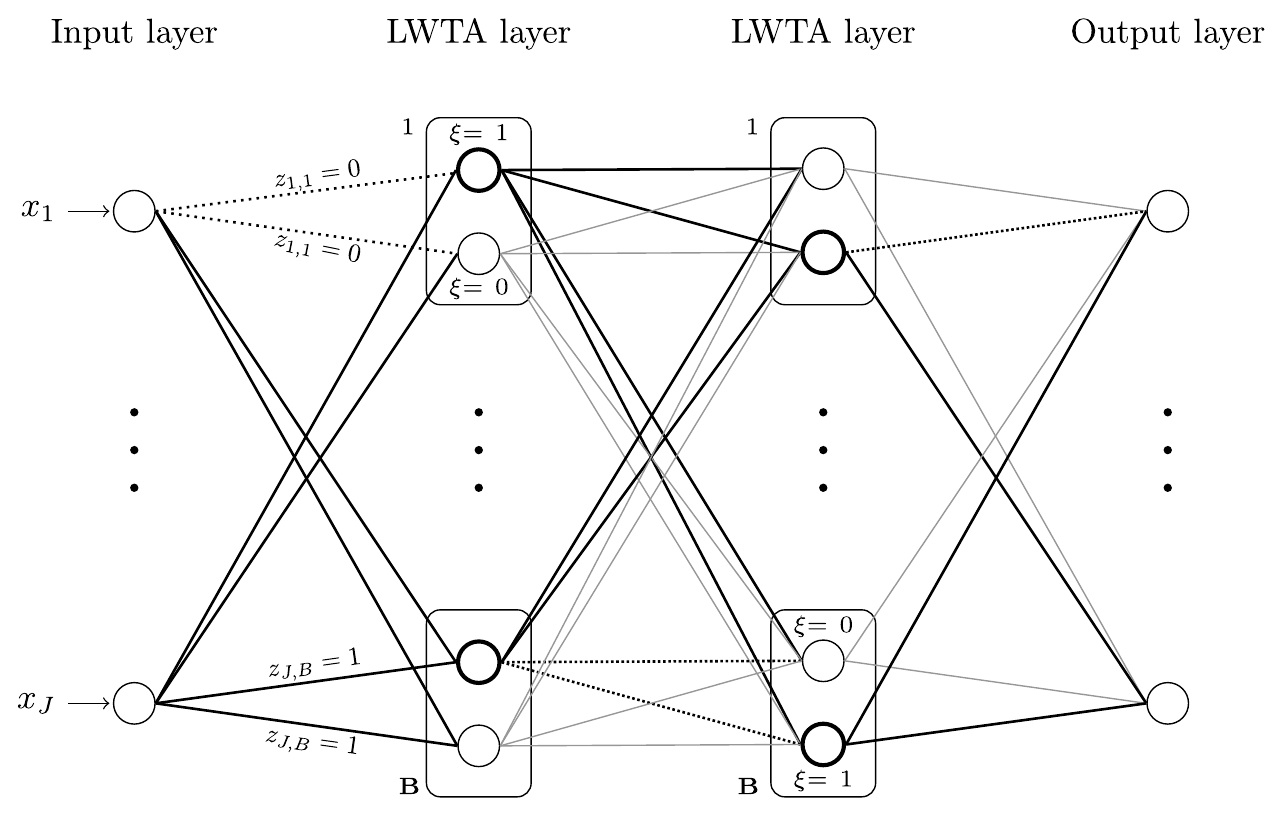}
			\caption{}
			\label{synopsis:wta}
		\end{subfigure}
		\begin{subfigure}[t]{.54\textwidth}
			\centering
			\includegraphics[scale=.55]{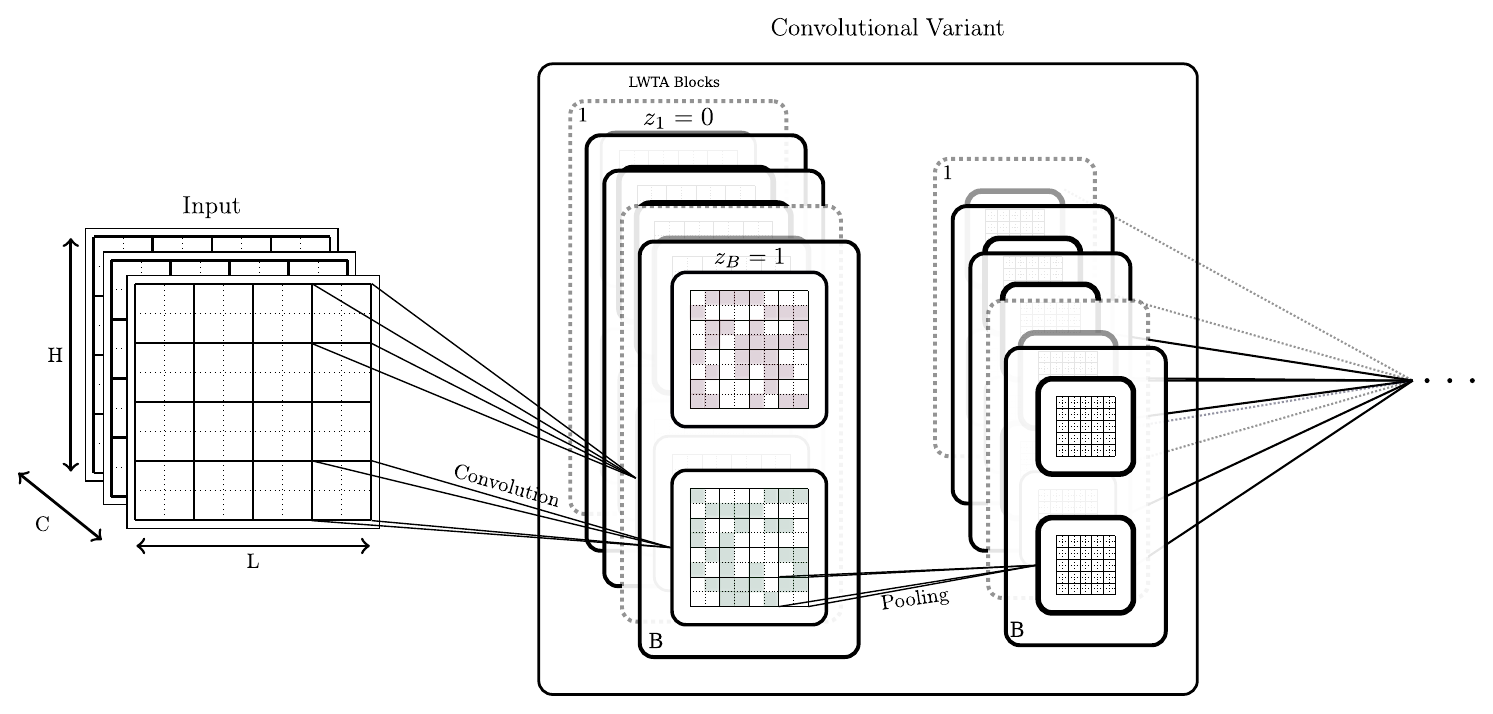}
			\caption{}
			\label{synopsis:fig:cnn_sb_lwta}
		\end{subfigure}
		\caption{ (a) A graphical representation of our competition-based modeling approach. Rectangles denote LWTA blocks, and circles the competing units therein. The winner units are denoted with bold contours ($\xi=1$). Bold edges denote retained connections ($z=1$). (b) The convolutional LWTA variant. Competition takes place among feature maps on a position-wise basis. The winner feature map at each position passes its output to the next layer, while the rest pass zero values at said position. }
	\end{figure*}

	We postulate that the winner unit latent indicators $\boldsymbol \xi_b, \forall b$ in Eq. \eqref{eqn:dense_y} are obtained via a \textit{competitive} random sampling procedure; this, translates to drawing samples from a Categorical distribution, where the probabilities are proportional to the intermediate linear computation that each unit performs.  Accordingly, the higher the linear response of a particular unit in a particular block, the higher its probability of being the winner in said block; this yields:
	{\small
		\begin{align}
			q(\boldsymbol \xi_b) = \mathrm{Categorical}\left( \boldsymbol \xi_b \Big| \mathrm{softmax}\left( \sum_{j=1}^J z_{j,b}\cdot [w_{j,b,u}]_{u=1}^U \cdot x_j \right)\right)
			\label{eqn:xi_dense}
		\end{align}
	}
	where $[w_{j,b,u}]_{u=1}^U$ denotes the vector concatenation of the set $\{w_{j,b,u}\}_{u=1}^U$. On the other hand, we postulate that the binary latent indicators $\boldsymbol Z$ are drawn from a Bernoulli distribution, operating in an ``on''-``off'' fashion, such that:
	\begin{align}
		q(z_{j,b}) = \mathrm{Bernoulli}(z_{j,b}|\tilde{\pi}_{j,b})
		\label{eqn:z_dense}
	\end{align}
	where $\tilde{\pi}_{j,b}, \ \forall j,b$ are trainable parameters. A graphical illustration of the proposed stochastic LWTA block is depicted in Fig.\ref{fig:block}. Each stochastic LWTA layer comprises multiple such LWTA blocks as illustrated in Fig. \ref{synopsis:wta}. At this point, it is important to note a key aspect of the proposed approach; that is, \textit{stochasticity}. In each layer, different stochastic representations arise due to the sampling procedure for both latent indicators $\boldsymbol \xi$ and $\boldsymbol Z$. Even when presented with the same input, different \textit{subnetworks} may be activated and different \textit{subpaths} are followed from the input to the output, as a result of winner, $\boldsymbol{\xi}$, and component utility, $\boldsymbol{Z}$, sampling.
	
	\subsection{A Convolutional Variant}
	
	To account for networks employing the convolutional operation, popular in RL, we formulate a convolutional variant of the proposed Stochastic LWTA rationale. Let us now consider an input tensor $\boldsymbol X \in \mathbb{R}^{H \times L \times C}$, where $H, L, C$ are the height, length and channels of the input. In this context, we define a set of kernels, each with weights $\boldsymbol W_b \in \mathbb{R}^{h \times l \times C \times U}$, where $h,l,C, U$ are the kernel height, length, channels and number of \textit{competing feature maps}; each layer comprises $B$ kernels. Analogously to the grouping and competition of linear units in dense layers, in this case, local competition is performed on a \textit{position-wise} basis among feature maps. Each kernel is treated as an LWTA block with \textit{competing feature maps}. The feature maps of each kernel compete to win the activation pertaining to each position; thus, we have as many competition outcomes within the kernel as the number of positions (in the definition of the feature maps). This way,  the latent winner indicator variables now pertain to the selection of winner feature maps on a \textit{position-wise} basis.
	
	Further, we introduce analogous latent utility indicators $\boldsymbol z \in \{0,1\}^B$ in order to infer which \textit{kernels} (LWTA blocks) are necessary for modeling the available data. Thus, here, if $z_b =0$, we omit \textit{whole blocks of competing feature maps}. Under this regard, each feature map $u=1, \dots, U$ in the $b$\textsuperscript{th} LWTA block (kernel) of a convolutional LWTA layer computes:
	\begin{align}
		\boldsymbol H_{b,u} = \left( z_b \cdot \boldsymbol W_{b,u} \right)\star \boldsymbol X \in \mathbb{R}^{H \times L}
		\label{eqn:h_conv}
	\end{align}
	Then, competition takes place among the $U$ kernel feature maps for claiming the available \textit{positions}, one by one. Specifically, the competitive random sampling procedure reads:
	{\small
		\begin{align}
			q(\boldsymbol \xi_{b,h',l'}) = \mathrm{Categorical}\left(\boldsymbol \xi_{b,h',l'} \Big | \mathrm{softmax}\left([\boldsymbol H_{b,u,h', l'}]_{u=1}^{U} \right) \right)
			\label{eqn:xi_conv}
		\end{align}
	}
	where $[\boldsymbol H_{b,u,h', l'}]_{u=1}^{U}$ denotes the concatenation of the set $\{\boldsymbol H_{b,u,h', l'}\}_{u=1}^{U}$.  In each kernel $b=1, \dots, B$, and for each position $h'=1, \dots, H$,  $l'=1, \dots, L$, only the winner feature map contains a non-zero entry; all the rest feature maps contain zero values at these positions. This yields sparse feature maps with mutually exclusive active pixels. Accordingly for the utility indicators:
	\begin{align}
		q(z_b) = \mathrm{Bernoulli}(z_b|\tilde{\pi}_b), \forall b
		\label{eqn:z_post}
	\end{align}
	\begin{figure*}
		\includegraphics[scale=0.65]{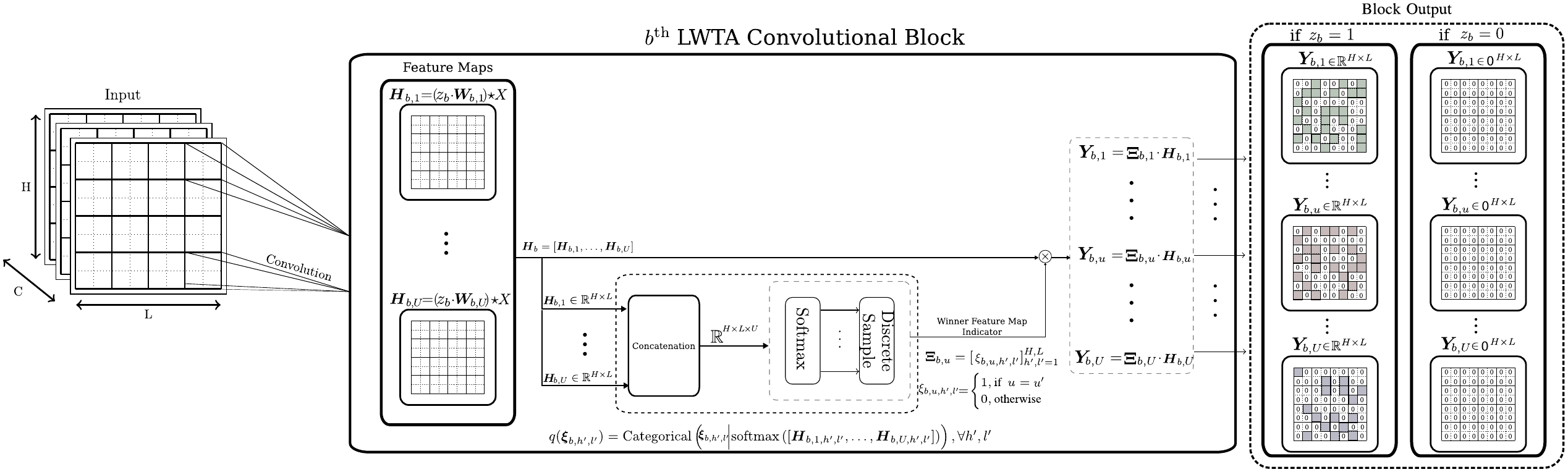}
		\caption{A detailed bisection of the $b$\textsuperscript{th} convolutional stochastic LWTA block. Presented with an input $\boldsymbol X \in \mathbb{R}^{H \times L \times C}$, competition now takes place among feature maps on a position-wise basis. Only the winner feature map contains a non-zero entry in a specific position. This leads to sparse feature maps, each comprising uniquely position-wise activated pixels. The component utility indicators $z_b$ encode which kernels/blocks the algorithm deems useful; when $z_b=0$, whole blocks are omitted.
		}
		\label{fig:block_conv}
	\end{figure*}
	Thus, the output $\boldsymbol Y \in \mathbb{R}^{H \times L \times B \cdot U}$ of a convolutional layer of the proposed stochastic LWTA-based networks is obtained via concatenation of the subtensors $\boldsymbol Y_{b,u}$ that read:
	\begin{align}
		\boldsymbol Y_{b,u} = \boldsymbol \Xi_{b,u} \left(\left( z_b \cdot \boldsymbol W_{b,u}\right) \star \boldsymbol X \right), \ \forall b,u
	\end{align}
	where $\boldsymbol \Xi_{b,u} = [\xi_{b,u,h',l'}]_{h',l'=1}^{H,L}$. The corresponding illustration of the proposed stochastic convolutional LWTA block is depicted in Fig. \ref{fig:block_conv}. Convolutional stochastic LWTA-based layers comprise multiple such blocks, as shown in  Fig. \ref{synopsis:fig:cnn_sb_lwta}.

	\section{Representation Leaning \& Mutual Information}
	\label{sec:background}
	
	Mutual information (MI) has been an indispensable tool in RL \cite{bell, linkser, contrastive}. In this context, \cite{tishby99information} introduced the Information Bottleneck (IB); this employs an information-theoretic objective that constrains the MI between the input and an intermediate representation, while enhancing the MI between said representation and the output \cite{tishby15, ShwartzZiv2017OpeningTB, michael2018on, dai2018compressing}. 
	
	In this line of work, the recently proposed Deep Variational Information Bottleneck (VIB) \cite{alemi2017} constitutes a variational approximation to the IB objective, yielding networks with increased generalization and robustness capabilities. Notably, there exists a connection between VIB and the well-known $\beta$-VAE formulation \cite{alemi2018uncertainty, alemi18a}. Both are founded on information theoretic arguments but used in different contexts; the first for supervised while the latter for unsupervised learning \cite{higgins2017beta-vae}. Here, we focus on the former. 
	
	Despite significant progress in the field, the majority of existing methods optimize a single information constraint to learn ``useful'' and diverse representations; however, this mode of operation fails to promote  diversity among the representations inferred from the latent units.
	
	Thus, various learning schemes have been devised in the literature to address this inadequacy; these are based on the idea of collaboration and competition between different neural representations \cite{tagger, neural_expectation, competitive_colab}. On this basis, \cite{icp} recently proposed the \textit{information-competing process} (ICP) scheme; this construct entails different representation parts that not only \textit{compete}, but also \textit{collaborate} to achieve a downstream task, thus enabling the model to learn richer and more discriminative representations. 
	
	\subsection{Proposed Approach}
	
	Let us denote by $t$, the output of the \textit{downstream task}, and by $\boldsymbol r$ the learned representation of $\boldsymbol x$; in supervised learning, $t$ corresponds to the target label pertaining to an observation $\boldsymbol x$. 
	Most information-theoretic approaches aim to maximize a single constraint, e.g., the MI between the representation $\boldsymbol r$ and target $t$ denoted as $\mathcal{I}(\boldsymbol r,t)$. In contrast, ICP takes a different route. Initially, in order to promote diversification of the emerging representations, $\boldsymbol r$ is \textit{explicitly split} into two different parts, $\{\boldsymbol \zeta, \boldsymbol y\}$; each is imposed different constraints: the ``information capacity'' of $\boldsymbol \zeta$ is minimized, while the ``information capacity'' of $\boldsymbol y$ is maximized. Under this regard, the objective reads:
	\begin{align}
		\max \left[ \mathcal{I}(\boldsymbol r, t) + \alpha \mathcal{I}( \boldsymbol y, \boldsymbol x) - \beta \mathcal{I}(\boldsymbol \zeta, \boldsymbol x) \right]
	\end{align}
	where $\alpha, \beta$ are regularization constants. $\mathcal{I}(\boldsymbol{\zeta}, \boldsymbol{x})$ and $\mathcal{I}(\boldsymbol{y},\boldsymbol{x})$ are called the \textit{mutual information minimization} and \textit{maximization} terms respectively.
	At this point, it is important to note that computing the MI between different pairs of variables  is usually intractable; thus, we need to devise different optimization schemes for each considered term. For example, to minimize $\mathcal{I}(\boldsymbol \zeta, \boldsymbol x)$, we introduce a \textit{variational approximation} $Q(\boldsymbol\zeta)$ to $P(\boldsymbol\zeta)$, yielding the following tractable upper bound:
	\begin{align}
		\mathcal{I}(\boldsymbol \zeta, \boldsymbol x ) \leq \mathbb{E}_{x\sim p(x)}\Big[ \mathrm{KL}[P(\boldsymbol\zeta|\boldsymbol x) || Q(\boldsymbol\zeta)]\Big]
	\end{align}
	where $Q(\boldsymbol{\zeta})$ is a standard Gaussian distribution. The parameters of this distribution, i.e., $\boldsymbol\mu_\zeta, \boldsymbol \sigma_\zeta$, are founded on an amortization scheme, similar to the well-known VAE formulation \citep{kingma2014autoencoding}; hence, deep neural networks are employed for their estimation.  
	
	In this context, before proceeding to define the full ICP objective, we \textit{highlight} that, \textit{all components} stemming from the corresponding tractable bounds and optimization schemes described next, e.g., feature extractors, discriminators and classifiers are originally implemented in \citep{icp} via ReLU-based DNNs. On this basis, this work introduces a radically different view: We employ deep networks built of novel \emph{stochastic LWTA} arguments, instead of standard non-linearities. Further, we complement this unique latent unit operation with a sparsity-inducing framework that allows determining the best postulated sub-network configuration.
	
	As already discussed, a notable aspect of ICP is the introduction of auxiliary constraints in order to prevent either part dominating the downstream task. To this end, the separated parts are individually allowed to accomplish $t$ via dedicated MI constraints, namely $\mathcal{I}(\boldsymbol \zeta,t)$ and $\mathcal{I}(\boldsymbol y, t)$; however, $\boldsymbol \zeta$ and $\boldsymbol y$ are \textit{constrained from knowing what each other has learned}. This property is realized via an additional constraint, that is by minimizing $\mathcal{I}(\boldsymbol \zeta, \boldsymbol y)$; this forces  $\boldsymbol \zeta$ and $\boldsymbol y$ to be independent of each other. The so-obtained objective reads:
	\begin{align}
		\begin{split}
			\mathcal{L}_{\text{ICP}}=\max \Big[ &\mathcal{I}(\boldsymbol{r},t) + \alpha \mathcal{I}(\boldsymbol{y}, \boldsymbol{x}) - \beta \mathcal{I}(\boldsymbol{x},\boldsymbol{\zeta})   \\ &+\mathcal{I}(\boldsymbol{\zeta},t) 
			+ \mathcal{I}(\boldsymbol{y},t) - \gamma \mathcal{I}(\boldsymbol{\zeta},\boldsymbol{y})\Big]
		\end{split}
		\label{eqn:icp_obj}
	\end{align}
	where $\gamma$ is another regularization constant, $\mathcal{I}(\boldsymbol{\zeta},t)$,  $\mathcal{I}(\boldsymbol{y},t)$, $\mathcal{I}(\boldsymbol{r},t)$ are called the \textit{inference} terms, and $\mathcal{I}(\boldsymbol{\zeta},\boldsymbol{y})$ the \textit{predictability minimization} term. $\mathcal{I}(\boldsymbol r, t)$ denotes the ``\textit{synergy}''  between the separate parts, aiming to accomplish the downstream task in a \textit{synergistic way}. Contrarily, the last three terms, i.e.,  $\mathcal{I}(\boldsymbol{\zeta},t)$, $\mathcal{I}(\boldsymbol{\zeta},\boldsymbol{y})$,  $\mathcal{I}(\boldsymbol{y},t)$ constitute the ``\textit{competition}'' aspect of the approach. This conception allows for both competition as well as synergy of the different representation parts, enhancing the information carried by said representations. In the following, we briefly present the required optimization schemes for each term of the objective.
	In the case of $\mathcal{I}(\boldsymbol y, \boldsymbol x)$ the KL divergence is divergent \citep{icp}; thus, we resort to maximization of the Jensen-Shannon (JS) divergence. Its variational estimation yields:
	\begin{align}
		\begin{split}
			\mathrm{JS}[&P(\boldsymbol y| \boldsymbol x)P(\boldsymbol x) || P(\boldsymbol y)P(\boldsymbol x))] = \\
			&\max \Big[ \mathbb{E}_{(y,x) \sim P(y|x)p(x)} \big[\log D(\boldsymbol y,\boldsymbol x)\big]\\
			&+ \mathbb{E}_{(\hat{y},x) \sim P(y)p(x)} \big[\log (1-D(\hat{\boldsymbol y},\boldsymbol x)\big]\Big]
		\end{split}
	\end{align}
	where $D(\cdot)$ is a discriminator, estimating the probability of an input pair; $(\boldsymbol y, \boldsymbol x)$ is the positive pair sampled from $P(\boldsymbol y|\boldsymbol x)P(\boldsymbol x)$ and $(\hat{\boldsymbol y}, \boldsymbol{x})$ is the negative pair sampled from $P(\boldsymbol y)P(\boldsymbol x)$; $\hat{\boldsymbol y}$ is a ``disorganized'' version of $\boldsymbol y$ \citep{icp}. For the \textit{inference} term $\mathcal{I}(\boldsymbol{r},t)$, the following lower bound is derived:
	{\small
		\begin{align}
			\mathcal{I}(\boldsymbol{r}, t) \geq \mathbb{E}_{x \sim P(x)}\Big[ \mathbb{E}_{r \sim P(r|x)}\Big[ \displaystyle\int P(t|x) \log Q(t|r) \ dt \Big] \Big]
			\label{eqn:inference_term}
		\end{align}
	}
	where $Q(t|\boldsymbol r)$ is a variational approximation of $P(t|\boldsymbol r)$, and $P(t|\boldsymbol x)$ denotes the distribution of the labels. Thus, Eq.\eqref{eqn:inference_term} essentially constitutes the cross-entropy loss. The expressions for the inference terms $\mathcal{I}(\boldsymbol{\zeta},t)$ and $\mathcal{I}(\boldsymbol{y},t)$ are similar. 
	
\begin{figure*}
		\centering
		\includegraphics[scale=0.4]{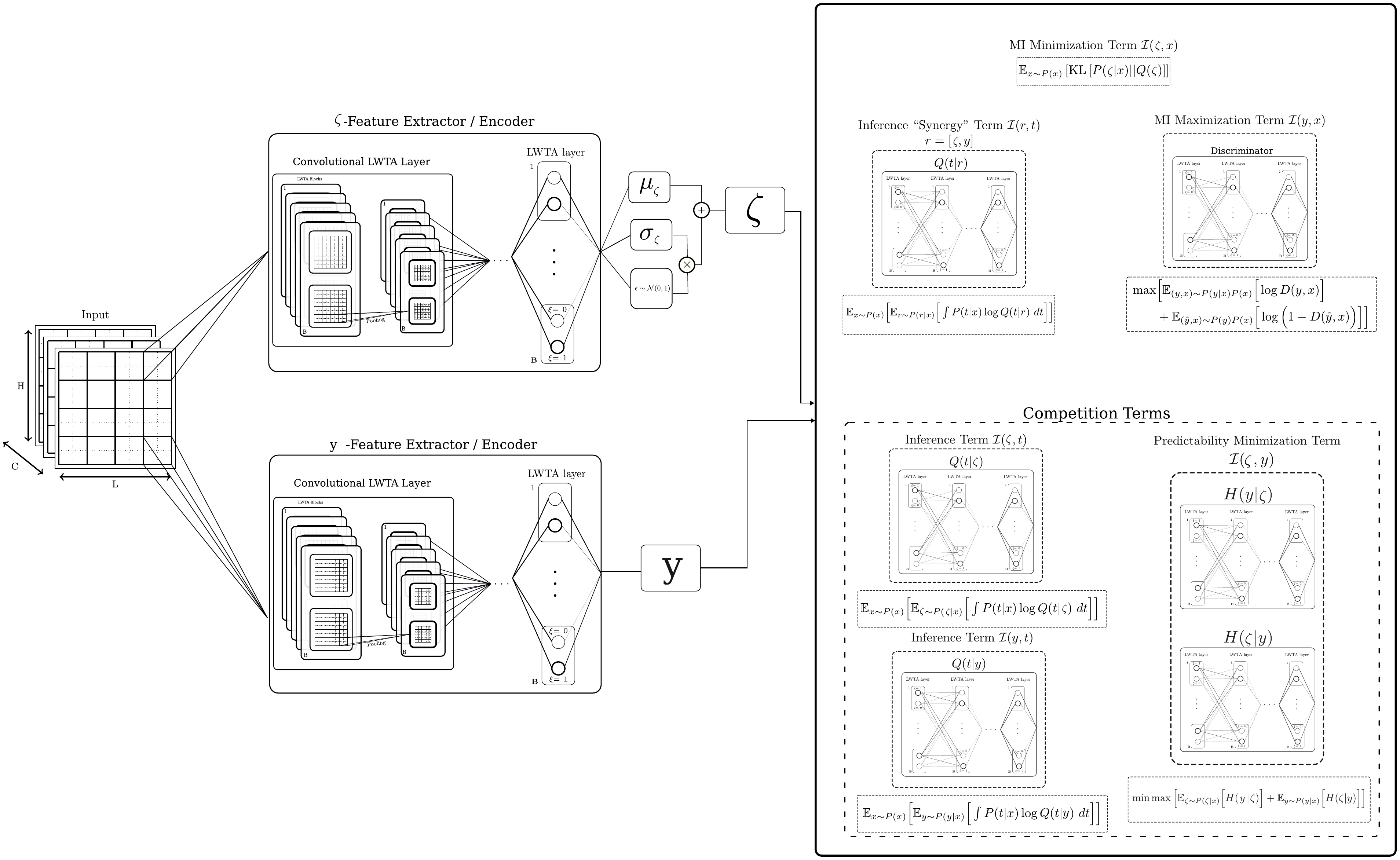}
		\caption{The ICP pipeline: The representation $\boldsymbol r$ is split into two parts $[\boldsymbol \zeta, \boldsymbol y]$. These, not only cooperate but also compete to accomplish the downstream task. This leads to the computation of the MI between different pairs of variables; this is usually intractable and optimized through different schemes. All components of these, are implemented via LWTA and IBP-based DNNs.}
		\label{fig:icp_lwta}
	\end{figure*}

	Finally, we turn to the \textit{predictability minimization term}; this enforces the representations parts to be independent. To this end, we introduce a predictor $H$ that is used to predict $\boldsymbol y$ given $\boldsymbol \zeta$; this way, we guide the $\boldsymbol \zeta$ variable extractor from producing $\boldsymbol \zeta$ values that can predict $\boldsymbol y$ \citep{icp}. The same procedure is followed for $\boldsymbol y$ to $\boldsymbol \zeta$. This yields:
	{\small
		\begin{align}
			\min\max \Big[ \mathbb{E}_{\zeta \sim P(\zeta|x)}[ H(\boldsymbol y|\boldsymbol\zeta)] + \mathbb{E}_{y\sim P(y|x)}[H(\boldsymbol\zeta|\boldsymbol y)]\Big]
		\end{align}
	}
	\noindent A graphical illustration of the described optimization process of the ICP objective of Eq.\eqref{eqn:icp_obj} is depicted in Fig. \ref{fig:icp_lwta}. Therein, all DNN-based components are founded on the Stochastic LWTA and component utility arguments of Section \ref{sec:lwta}.
	\subsection{Training \& Prediction}	
	\label{sec:training}
	
	The core training objective was defined in the previous section; this stems from the ICP rationale as expressed in Eq.\eqref{eqn:icp_obj}; however,  the existence of Stochastic LWTA activations and the component utility mechanism, necessitates the augmentation of the final objective via appropriate KL terms.
	
	Without loss of generality, we begin by considering a symmetric Categorical distribution for the latent variable indicators $\boldsymbol \xi$; hence, $p(\boldsymbol \xi_b) = \mathrm{Categorical}(1/U) \forall b$ for the dense layers, and $p(\boldsymbol \xi_{b,h',l'}) = \mathrm{Categorical}(1/U), \ \forall b,h',l'$ for convolutional ones. Differently, for the latent utility indicators $\boldsymbol Z$ (or $\boldsymbol z)$ we do not impose a symmetric prior; instead, we turn to the non-parametric Bayesian framework and specifically to the Indian Buffet Process (IBP) \citep{ghahramani2006infinite}. This constitutes a sparsity-promoting prior; at the same time, its so-called \textit{stick-breaking process} (SBP) \citep{teh2007stick} renders IBP amenable to Variational Inference. The hierarchical construction reads: 
	{\small
		\begin{align}
			p(z_{j,b}) = \mathrm{Bernoulli}(\pi_b), \ \pi_b = \prod_{i=1}^b u_b, \ u_b \sim \mathrm{Beta}(\omega, 1 )
			\label{eqn:ibp_prior}
		\end{align}
	}
	where $\omega$ is a non-negative constant, controlling the induced sparsity.  The SBP requires an additional set of latent \textit{stick} variables $u_b, \forall b$. Since these are $\mathrm{Beta}$-distributed, we assume a posterior of similar form: $q(u_b) = \mathrm{Beta}(u_b | \tilde{a}_b, \tilde{b}_b)$, where $\tilde{a}_b, \tilde{b}_b, \forall b$ are trainable variational parameters. This yields:
	\begin{align}
		\begin{split}
			\mathcal{L} = \mathcal{L}_{\text{ICP}} - \mathrm{KL}\big[ q(\boldsymbol \xi) || p(\boldsymbol \xi)\big] &- \mathrm{KL} \big[ q(\boldsymbol Z) || p(\boldsymbol Z) \big] \\
			&- \mathrm{KL} \big[ q(\boldsymbol u) || p(\boldsymbol u) \big]
		\end{split}
		\label{eqn:final_obj}
	\end{align}
	For training, we perform Monte-Carlo sampling to estimate Eq. \eqref{eqn:final_obj} using a single reparameterized sample for each latent variable. For the Gaussian distributed variables, e.g. $\boldsymbol\zeta$, we resort to the well-known Gaussian trick. For $\boldsymbol \xi$ and $\boldsymbol Z$, these are obtained via the continuous relaxation of the Categorical and Bernoulli distribution \citep{jang, maddison}. In the following, we describe the reparameterization trick for the $\boldsymbol\xi$ variables of dense layers; the cases for $\boldsymbol Z$ and for the convolutional variant are analogous.
	
	Let $\tilde{\boldsymbol\xi}$ denote the probabilities of $q(\boldsymbol \xi)$ (Eqs.\eqref{eqn:xi_dense},\eqref{eqn:xi_conv}). Then, the samples $\hat{\boldsymbol\xi}$ can be expressed as:
	\begin{align}
		\hat{\xi}_{b,u} = \mathrm{Softmax}((\log \tilde{\xi}_{b,u} + g_{b,u})/\tau), \forall b,u
	\end{align}
	where $g_{b,u} = -\log(-\log V_{b,u}), \ V_{b,u}\sim \mathrm{Uniform}(0,1)$ and $\tau \in (0, \infty)$ is a \textit{temperature} constant that controls the degree of the approximation. Similarly, the Beta distribution of the stick variables $\boldsymbol u$ is not readily amenable to reparameterization; for these variables, we obtain the required samples via the Kumaraswamy distribution \citep{kumaraswamy}; this constitutes an approximation of Beta and admits the following reparameterization trick:
	\begin{align}
		\hat{u}_b = \left( 1 - \left( 1- G\right)^{\frac{1}{\tilde{a}_b}}\right)^{\frac{1}{\tilde{b}_b}}
	\end{align}
	where $G \sim \mathrm{Uniform}(0,1)$ and $\tilde{a}_b, \tilde{b}_b$ are the variational parameters of the original Beta distribution. We can now compute each expectation term in the objective (Eq.\eqref{eqn:final_obj}) via these samples. For example, we can write the KL divergence term for the latent variables $\boldsymbol \xi$ as:
	\begin{align}
		\begin{split}
			\mathrm{KL}\big[ q(\boldsymbol\xi_b) || p(\boldsymbol\xi_b) \big]&= \mathbb{E}_{q(\boldsymbol \xi_b)}\big[\log q(\boldsymbol\xi_b) - \log p(\boldsymbol\xi_b) \big]\\
			&\approx \log q(\hat{\boldsymbol\xi}_b) - \log p(\hat{\boldsymbol\xi}_b), \ \forall b
		\end{split}
	\end{align}	
	
	At prediction time, we directly draw samples from the trained posteriors $q(\boldsymbol\xi)$ and $q(\boldsymbol z)$ in order to determine the winner in each block of the network and to assess component utility respectively. Thus, each time we sample, even for the same input, a different \textit{subpath} may be followed according to the outcomes of the sampling processes. This leads to a  stochastic alternation of the emerging representations of the network at each forward pass.
	
	\section{Experimental Evaluation}
	\label{sec:experimental}
	We evaluate our model on image classification, where we consider two popular benchmarks, namely CIFAR-10 and CIFAR-100 \cite{Krizhevsky09learningmultiple}, containing natural images with 10 and 100 classes respectively. We compare our approach to recent information-theoretic approaches to deep networks, including VIB \cite{alemi2017}, DIM \cite{dim} and ICP \cite{icp}. To this end, we follow the same experimental setup as \cite{icp}, and consider four different networks for both datasets: (i) VGG-16 \cite{vgg}, (ii) GoogLeNet \cite{googlenet}, (iii) ResNet20 \cite{resnet}, and (iv) DenseNet40 \cite{densenet}. For both datasets, we normalize the raw image values using the per-channel means and standard deviations. We employ some common data augmentation procedures, including random cropping and mirroring.

	\subsection{Experimental Setup}
	
	To evaluate our approach, we consider two different setups for our competition-based LWTA activations: (i) an architecture comprising LWTA blocks with $U=2$ competing units/feature maps, denoted as ICP\textsubscript{LWTA-2}; and (ii) an architecture comprising LWTA blocks with $U=4$ competing units/feature maps, denoted as ICP\textsubscript{LWTA-4}. In all cases, the total number of hidden units/features in each layer remain the same as in the original ReLU-based architectures. This ensures comparability (size-wise) among existing approaches.
	
	We also compare performance when we employ the proposed, IBP-driven, network sampling mechanism, and when we omit it. For clarity, in the following, we denote our full model (which employs the IBP-based mechanism) as ICP\textsubscript{IBP \& LWTA-2} and ICP\textsubscript{IBP \& LWTA-4}, respectively. In this context, to allow for compressing the model by exploiting the trained posteriors over the latent indicators $z$, we adopt the following rationale: After network training, we introduce a cut-off threshold $\tau = 0.001$. All components with trained utility posterior $\tilde{\pi} \triangleq q(z=1) < \tau$ are removed from the network; all rest are retained and used at prediction time. In the following tables, the \textit{compression} metric corresponds to the ratio of the number of network components removed to the total number of network components.
	
	We choose an uninformative Beta prior for the IBP (Eq.\eqref{eqn:ibp_prior}): $\mathrm{Beta}(1,1)$; thus, $\omega =1$. For the Gumbel-Softmax relaxation, we set the temperatures $\tau$ for the prior and posterior distributions to $0.5$ and $0.67$ respectively \cite{maddison}. Further initialization procedures and hyperparameter values can be found in the Supplementary.
	
	We draw a single (reparameterized) sample from all the involved random variables  during training, while we draw $5$ different samples during inference (and proceed with Bayesian averaging). We perform multiple evaluations for each architecture and dataset, i.e., 5 runs, and report the best performing one. All experiments were run on a workstation with 2x Quadro P6000 24GB GPUs and 64GB RAM.

	\subsection{Experimental Results}
	\begin{table}[h!]
		\centering
		\resizebox{\columnwidth}{!}{
			\begin{tabular}{l|c|c|c|c}
				\hline
				Model &  VGG-16  & GoogLeNet  & ResNet20& DenseNet40 \\\hline
				& \multicolumn{4}{c}{Error (\%) $\parallel$ Compression (\%)}\\\hline
				Baseline & 6.67 $\parallel$ 0.00  & 4.92 $\parallel$ 0.00 & 7.63 $\parallel$ 0.00 & 5.83 $\parallel$ 0.00 \\\hline
				VIB \cite{alemi2017} & 6.81 $\parallel$ 0.00  & 5.09 $\parallel$ 0.00  & 6.95 $\parallel$ 0.00  & 5.72 $\parallel$ 0.00 \\
				DIM*\cite{hjelm2018learning} & 6.54 $\parallel$ 0.00  & 4.65 $\parallel$ 0.00  & 7.61 $\parallel$ 0.00  & 6.15 $\parallel$ 0.00 \\
				VIB\textsubscript{$\times$2}  & 6.86 $\parallel$ 0.00 & 4.88 $\parallel$ 0.00  & 6.85 $\parallel$ 0.00  & 6.36 $\parallel$ 0.00 \\
				DIM*\textsubscript{$\times$2} & 7.24 $\parallel$ 0.00  & 4.95 $\parallel$ 0.00  & 7.46 $\parallel$ 0.00  & 5.60 $\parallel$ 0.00  \\
				ICP\textsubscript{-ALL} & 6.97 $\parallel$ 0.00  & 4.76 $\parallel$ 0.00  & 6.47 $\parallel$ 0.00  & 6.13 $\parallel$ 0.00  \\
				ICP\textsubscript{-COM} & 6.59 $\parallel$ 0.00 & 4.67 $\parallel$ 0.00 &  7.33 $\parallel$ 0.00  & 5.63 $\parallel$ 0.00  \\
				ICP & 6.10 $\parallel$ 0.00  & \textbf{4.26} $\parallel$ 0.00  & 6.01 $\parallel$ 0.00  & 4.99 $\parallel$ 0.00 \\\hline
				ICP\textsubscript{IBP \& LWTA-2} & \textbf{6.01} $\parallel$ \textbf{40.4}  & 4.31  $\parallel$ \textbf{35.2}  & \textbf{5.94}  $\parallel$ \textbf{37.1}  & \textbf{4.78}  $\parallel$ \textbf{32.0} \\
				ICP\textsubscript{IBP \& LWTA-4} & 7.02  $\parallel$ 30.4  & 4.74   $\parallel$ 28.2  & 6.30  $\parallel$ 31.3  & 5.61  $\parallel$ 26.3  \\
				\hline
			\end{tabular}
		}
		\caption{CIFAR-10.}
		\label{table:cifar10}
	\end{table}
	\textbf{CIFAR-10.} We train the considered four networks adopting the number of epochs and the optimization parameters used in the original ICP implementation \cite{icp}; this ensures transparency and comparability of the empirical results. The exact experimental setup for each network can be found in the Supplementary. The obtained comparative results for all considered methods and networks can be found in Table \ref{table:cifar10}. Therein, \textit{Baseline} corresponds to the original architectures without any type of MI constraints; VIB\textsubscript{$\times$2} and DIM*\textsubscript{$\times$2} denote the VIB and DIM* methods where the representation dimension has been expanded in order to facilitate a fair (size-wise) comparison with ICP. As we observe, our proposed method yields competitive (in many cases better) classification accuracy over the best-performing baseline ICP alternative. At the same time, by exploiting the IBP-based mechanism, it allows for powerful compression of the considered architecture; this is effected by removing all network components (and associated weights) with trained utility posteriors below the cut-off threshold. Clearly, the aim of this work is not to propose a method for deep network compression. We posit, though, that this compression process 
	further facilitates the diversification of the learned representations. We explore these aspects in Section 4.4. 
	
	\textbf{CIFAR-100.} For CIFAR-100, we follow an analogous procedure. The classification results are shown in Table \ref{table:cifar100}. In this set of experiments, the increased classification capabilities of the baseline ICP model, that is the conventional ICP model with ReLU-based nonlinearities, compared to the other information-theoretic approaches is more evident. Our proposed approach, once again follows this trend, yielding on-par or better classification accuracy with ICP, while at the same time exhibiting significant compression capabilities.
	\begin{table}[h!]
		\centering
		\resizebox{\columnwidth}{!}{
			\begin{tabular}{l|c|c|c|c}
				\hline
				Model &  VGG-16  & GoogLeNet  & ResNet20 & DenseNet40 \\\hline
				& \multicolumn{4}{c}{Error (\%) $\parallel$ Compression (\%)}  \\\hline
				Baseline & 26.41 $\parallel$ 0.00 & 20.68 $\parallel$ 0.00 & 31.91 $\parallel$ 0.00 & 27.55 $\parallel$ 0.00\\\hline
				VIB \cite{alemi2017} & 26.56 $\parallel$ 0.00 & 20.93 $\parallel$ 0.00 & 30.84 $\parallel$ 0.00 & 26.37 $\parallel$ 0.00 \\
				DIM*\cite{hjelm2018learning} & 26.74 $\parallel$ 0.00 & 20.94 $\parallel$ 0.00 & 32.62 $\parallel$ 0.00 & 27.51 $\parallel$ 0.00\\
				VIB\textsubscript{$\times$2}  & 26.08 $\parallel$ 0.00 & 22.09 $\parallel$ 0.00 & 29.74 $\parallel$ 0.00 & 29.33 $\parallel$ 0.00\\
				DIM*\textsubscript{$\times$2} & 25.72 $\parallel$ 0.00 & 21.74 $\parallel$ 0.00 & 30.16 $\parallel$ 0.00 & 27.15 $\parallel$ 0.00\\
				ICP\textsubscript{-ALL} & 26.73 $\parallel$ 0.00 & 20.90 $\parallel$ 0.00 & 28.35 $\parallel$ 0.00 & 27.51 $\parallel$ 0.00 \\
				ICP\textsubscript{-COM} & 26.37 $\parallel$ 0.00 & 20.81 $\parallel$ 0.00 &  32.76 $\parallel$ 0.00 & 26.85 $\parallel$ 0.00 \\
				ICP & 24.54 $\parallel$ 0.00 & \textbf{18.55} $\parallel$ 0.00 & 28.13 $\parallel$ 0.00 & 24.52 $\parallel$ 0.00\\\hline
				ICP\textsubscript{IBP \& LWTA-2} & \textbf{24.35}   $\parallel$ \textbf{32.0} & 19.00  $\parallel$ \textbf{29.0} &   \textbf{28.02} $\parallel$  \textbf{31.5} & \textbf{24.44}  $\parallel$ \textbf{29.2}\\
				ICP\textsubscript{IBP \& LWTA-4} & 25.44  $\parallel$ 22.0 & 20.12  $\parallel$ 26.5 & 29.34  $\parallel$ 21.2 & 25.07  $\parallel$ 24.4\\
				\hline
			\end{tabular}
		}
		\caption{CIFAR-100.}
		\label{table:cifar100}
	\end{table}

	\textbf{Random Seed Effect.} A core aspect of the considered approach lies in its stochastic nature. Thus, in order to assess the overall robustness of the proposed method, we run each experiment 5 times and report the means and standard deviations; these are presented in Table \ref{table:means}. We observe that our approach exhibits consistent performance in all cases, while in most occasions, the mean performance obtained by multiple runs outperforms the baseline ICP approach.
	\begin{table}[h!]
		\centering
		\resizebox{\columnwidth}{!}{
			\begin{tabular}{l|c|c|c|c}
				\hline
				\multicolumn{5}{c}{CIFAR-10}\\\hline
				& \multicolumn{4}{c}{Mean (\%) $\parallel$ Standard Deviation (\%)}  \\\hline
				Model &  VGG-16  & GoogLeNet  & ResNet20 & DenseNet40 \\\hline
				
				ICP\textsubscript{IBP \& LWTA-2} & 6.07 $\parallel 0.04$  & 4.35 $\parallel 0.04$   & 6.00 $\parallel 0.06$  & 4.87 $\parallel 0.05$   \\
				ICP\textsubscript{IBP \& LWTA-4} & 7.12 $\parallel 0.08$  & 4.84 $\parallel 0.06$ & 6.42 $\parallel 0.10$  & 5.70 $\parallel 0.08$   \\
				\hline\hline 
				
				\multicolumn{5}{c}{CIFAR-100}\\\hline
				& \multicolumn{4}{c}{Mean (\%) $\parallel$ Standard Deviation (\%)}  \\\hline
				Model &  VGG-16  & GoogLeNet  & ResNet20 & DenseNet40 \\\hline
				
				ICP\textsubscript{IBP \& LWTA-2} & 24.52 $\parallel 0.12$  & 19.25 $\parallel 0.14$   &  28.10 $\parallel 0.05$ & 24.52 $\parallel 0.05$\\
				ICP\textsubscript{IBP \& LWTA-4} & 25.59 $\parallel 0.08$ & 20.28 $\parallel 0.10$   & 29.51 $\parallel 0.10$   & 25.19 $\parallel 0.08$  \\\hline
			\end{tabular}
		}
		\caption{Means and standard deviations of 5 different runs under different seeds for all datasets and architectures.}
		\label{table:means}
	\end{table}
	
	\subsection{Ablation Study}
	
	Here, we focus on the VGG-16 architecture described in the previous section, and assess the individual impact of each of the proposed components, i.e. LWTA and IBP, to the classification performance of the network. At the same time, we examine whether adoption of a deterministic LWTA scheme, as opposed to the adopted stochastic construction, would yield equal or inferior performance in these benchmarks. 
	
	The obtained comparative results are depicted in Table \ref{table:ablation}. Therein, the ICP\textsubscript{LWTA-*\textsuperscript{max}} and ICP\textsubscript{IBP \& LWTA-*\textsuperscript{max}} entries correspond to networks where the LWTA competition function picks the unit with greatest activation value, in a deterministic fashion; $*=2,4$ denotes the number of competitors in each LWTA block. We consider both omission and use of the IBP-based mechanism, respectively. 
	
	As we observe, for both $U=2,4$ settings, the Stochastic LWTA approach outperforms deterministic LWTA units picking the greatest value. The IBP-based mechanism seems to also facilitate classification accuracy (in addition to compressing the network we perform inference with).
	
	\begin{table}
		\centering
		\caption{Ablation Study: CIFAR-10 test-set using a VGG-16 \cite{vgg} architecture.} 
		\label{table:ablation}
		\begin{tabular}{l|c}
			\hline
			Model &  VGG-16 \\\hline
			Baseline & 6.67 \\
			ICP & 6.10 \\\hline
			ICP\textsubscript{LWTA-2\textsuperscript{max}} & 6.34\\
			ICP\textsubscript{LWTA-2} & 6.23 \\
			ICP\textsubscript{IBP \& LWTA-2\textsuperscript{max}} & 6.27 \\
			ICP\textsubscript{IBP \& LWTA-2} & \textbf{6.01}  \\\hline
			ICP\textsubscript{LWTA-4\textsuperscript{max}} &  7.01\\
			ICP\textsubscript{LWTA-4} & 6.85  \\
			ICP\textsubscript{IBP \& LWTA-4\textsuperscript{max}} & 7.32 \\
			ICP\textsubscript{IBP \& LWTA-4} & 7.02 \\
			\hline
		\end{tabular}
		
	\end{table}
	
	\subsection{Representation Diversification}

	The ultimate goal of this work is to allow for deep networks to yield representations that are sufficiently diverse. In this Section, we investigate the diversification capabilities of the proposed framework, both qualitatively and quantitatively. To perform this analysis, and due to space limitations, we focus on the VGG-16 architecture and the CIFAR-10 dataset. Similar results on further architectures and datasets can be found in the Appendix.

	\begin{figure}[h!]
		\centering
		\includegraphics[scale=0.95]{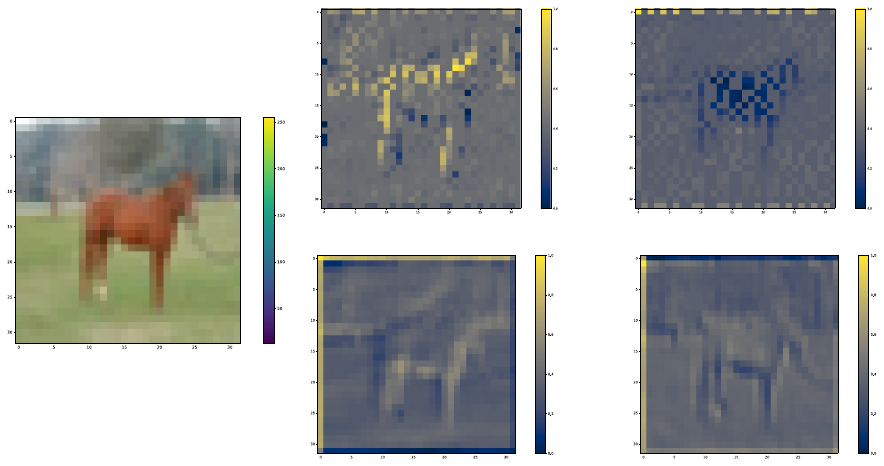}
		\caption{Feature map visualizations for a test example of the CIFAR-10 dataset, emerging from the first layer of the considered VGG-16 architecture. (Left) The original image, (Upper Row) a visualization of the outputs of the two competing feature maps of the first LWTA block ($U=2$), and (Bottom Row) a visualization of the first two filters of the conventional ReLU-based approach. The latter exhibit significant overlap, contrary to the much diverse representations of our approach which are mutually-exclusive.}
		\label{fig:acts7}
	\end{figure}

	To perform a qualitative evaluation, we visually compare the emerging \textit{intermediate representations from the hidden layers} of employed ReLU-based networks (standard ICP) and our approach. As we observe in Fig. \ref{fig:acts7}, there exists a clear disparity between the proposed approach and the commonly employed nonlinearities. Clearly, the ReLU-based architecture allows for more ``aesthetically pleasing'' representations. However, the proposed networks yield significant \textit{diversification}, as they split the resulting representations from each block to \textit{mutually exclusive parts}. This constitutes a radically different RL scheme with significant diversification capacity for the emerging \emph{intermediate} representations.
	%
	
	%
	For the quantification of the diversification properties, we turn to the commonly used Linear Separability metric. This metric constitutes a proxy for quantifying the disentanglement and MI between the emerging representations and the class labels \cite{dim}. In this context, linear classification is usually considered, and for this we use the standard Support Vector Machine (SVM) approach. To this end, we hold the encoders of $\boldsymbol{y}$ and $\boldsymbol{\zeta}$ fixed, and build separate SVM-based classifiers on the two representation parts $\boldsymbol \zeta$ , $\boldsymbol y$, their combination (denoted as \textrm{Total}), as well as the output of the last convolutional layer of the $\boldsymbol\zeta$ encoder (Conv), which stems either from ReLU units or LWTA blocks. 
	
	The obtained comparative separability results are depicted in Table \ref{table:svm}. We observe that, by training a linear SVM model on the obtained representations $\boldsymbol \zeta$ or $\boldsymbol y$, their combination thereof, or the last convolutional layer of the underlying encoder, the linear model can obtain more potent accuracy profile if its inputs stem from networks formulated under our approach (LWTA \& IBP arguments). This hints at obtaining more diversified representations, that therefore convey richer information to the linear SVM classifier, which is also easier to discern in a linear classification fashion.
	
	\begin{table}[h!]
		\centering
		\caption{Results on linear separability using SVMs.}
		\label{table:svm}
		\resizebox{0.95\columnwidth}{!}{
			\begin{tabular}{c|cccc}
				\hline
				\multirow{2}{*}{Method} & \multicolumn{4}{c}{Proxies}  \\
				& SVM($\boldsymbol\zeta$) & SVM($\boldsymbol y$) & SVM(\textrm{Total}) & SVM(Conv) \\\hline
				ICP & 91.5 & 91.9 & 92.9 & 31.2\\
				ICP\textsubscript{LWTA-2 \& IBP} & \textbf{91.9} & \textbf{92.4} & \textbf{93.4} & \textbf{32.2}\\\hline
			\end{tabular}
		}
		
	\end{table}

	\section{Conclusions}
	\label{sec:conclusions}
	
	In this work, we attacked the problem of promoting diversified representations in Deep Learning. To this end, we introduced principled network arguments formulated by stochastic competition-based Local Winner-Takes-All activations. We combined these with network component utility mechanisms, which rely on the use of the IBP prior. Then, to further enrich the emerging representations, we employed information-theoretic arguments, founded on competing mutual information constraints under the Information Competing Process. This results in an efficient network training and prediction scheme, that significantly compresses the networks during prediction. We performed a thorough experimental evaluation, using benchmark datasets and  several standard network architectures. We compared networks crafted using the proposed arguments against standard, ReLU-based constructions. Our experimental results provided strong empirical evidence of the efficacy of the proposed framework. Specifically, in all cases, our approach yielded on-par or improved accuracy for significantly compressed networks. At the same time,
	our qualitative and quantitative analysis of the obtained representations showed our approach results in representations that: (i) visually appear much more diverse; and (ii) are more informative to a linear classifier trained on them, specifically an SVM used as a proxy to linear separability.

	\section*{Acknowledgements}
	
	This work has received funding from the European Union's Horizon 2020 research and innovation program
	under grant agreement No 872139, project aiD.
	
	\bibliography{aaai22}

\end{document}